\documentclass[letterpaper]{article}

\usepackage[preprint]{aaai2027}
\usepackage[hyphens]{url}
\usepackage{graphicx}
\usepackage{natbib}
\usepackage{caption}
\usepackage{booktabs}
\usepackage{multirow}
\usepackage{array}
\usepackage{amsmath}
\usepackage{amssymb}
\usepackage{xcolor}
\usepackage{algorithm}
\usepackage{algorithmic}

\usepackage{amsmath,amsfonts,bm}

\def\eqref#1{equation~\ref{#1}}

\def\1{\bm{1}}

\DeclareMathAlphabet{\mathsfit}{\encodingdefault}{\sfdefault}{m}{sl}
\SetMathAlphabet{\mathsfit}{bold}{\encodingdefault}{\sfdefault}{bx}{n}

\title{Spec-AUF: Accept-Until-Fail Training under Train-Inference Misalignment for Masked Block Drafters}

\author{
    Tianjian Yang\textsuperscript{\rm 1},
    Meng Li\textsuperscript{\rm 2,3}\thanks{Corresponding author.}
}
\affiliations{
    \textsuperscript{\rm 1}School of Electronics Engineering and Computer Science, Peking University, Beijing, China\\
    \textsuperscript{\rm 2}Institute for Artificial Intelligence, Peking University, Beijing, China\\
    \textsuperscript{\rm 3}School of Integrated Circuits, Peking University, Beijing, China\\
    2300012738@stu.pku.edu.cn, meng.li@pku.edu.cn
}

\newcommand{\method}{AUF}

\begin{document}

\maketitle

\begin{abstract}
Speculative decoding accelerates autoregressive generation by drafting a block of
tokens that the target model verifies left-to-right, committing only the longest
accepted prefix. Block (DLM-style) drafters predict the whole block in parallel,
which is fast but trained with a full-block cross-entropy that supervises every
position against the gold continuation---even though inference discards every
token after the first rejection. Recent acceptance-aware objectives patch this by
reweighting the full-block loss; we instead use teacher-forced learning as a
motivation for how supervision should concentrate on the accepted prefix. A
mask-only block drafter has no input-side channel for gold-prefix conditioning,
so AUF approximates that prefix-sensitive supervision on the loss side by keeping
the cross-entropy support only through the drafter's first predicted failure.
AUF is a single, detached change to the CE support---no auxiliary objective, no
verifier rollouts, and no change to the inference pipeline or the exactness
contract. Within fixed drafter backbones and serving settings on Qwen3-8B, AUF
raises the DFlash drafter's average emitted length $\tau$, averaged over six
benchmarks, from $2.40$ to $2.61$, with a gain on every benchmark, and transfers
to Domino's two-branch head ($2.56$ to $2.68$). Two findings sharpen the picture:
the decay-only baseline reaches \emph{higher} token accuracy on the shared block
mask yet decodes \emph{worse}, and on DFlash, once AUF truncates the support, the
standard exponential position-decay weighting becomes empirically inert.
\end{abstract}

\section{Introduction}
\label{sec:introduction}

Speculative decoding (SD) accelerates autoregressive language-model inference by
letting a lightweight draft model propose several future tokens and letting the
target model verify them in one parallel pass~\citep{leviathan2023speculative,chen2023speculative}.
Because verification accepts only the longest target-consistent prefix, the
practical speedup of SD is governed less by isolated token accuracy than by the
draft model's ability to produce a consecutive prefix that the target will
accept. A drafter that predicts later tokens correctly after an early mistake
still wastes those predictions: the verifier rejects the suffix once the prefix
is broken.

This observation changes how draft-model training should be viewed. Standard
alignment objectives train the drafter to match target tokens over a full block,
often with a hand-designed position prior --- a fixed weight that decays with
block position, encoding the inductive bias that earlier tokens matter more
(DFlash, for instance, uses an exponential decay~\citep{chen2026dflash}). These
objectives recognize that early positions matter more, but they still supervise
tokens whose contribution to throughput is conditional on all previous draft
tokens being accepted. Draft-model training is therefore not only a
token-alignment problem; it must also reflect the left-to-right prefix contract
used by the verifier.

Our starting point is the analogy to autoregressive draft-model training. With
teacher-forced maximum likelihood, an autoregressive drafter receives the gold
prefix on its input side, so the loss at position $i$ is conditioned on the event
that positions $<i$ are already correct~\citep{li2025eagle3}. Gold supervision is
the strong signal, and the design question is which positions are entitled to
receive it. A masked or diffusion-style block drafter such as DFlash cannot obtain
this conditioning from its input at all: the block positions are predicted in
parallel from masked tokens. We therefore approximate that prefix-sensitive
supervision on the loss side. Accept-Until-Fail (\method) keeps the gold-target
cross-entropy only through the first draft error: the accepted prefix and the
first failing token remain active, while all later positions are masked out.

The closest related line begins with GRIFFIN, which also drops the draft loss
once an earlier target token falls outside the model's Top-$K$
predictions~\citep{hu2025griffin}; AUF shares this ``stop supervising once the
prefix is broken'' intuition but differs in both criterion and mechanism, while
the soft-weighting line (decay, D-PACE) keeps a full-block support and
re-approximates the conditioning with position weights
instead~\citep{wu2026dpace}. Section~\ref{sec:related} places these treatments of
the same training--verification mismatch on one axis; what we add is a
teacher-forcing-motivated reading and the mask-only setting it applies to, where
the prefix conditioning has no input-side channel and must act on the loss side.

This direction is not an obvious win. D-PACE reports that a GRIFFIN-style Top-3
prefix-mask baseline \emph{underperforms} even the decayed-CE DFlash baseline on
Qwen3-4B~\citep{wu2026dpace} --- but Top-3 \emph{membership} is GRIFFIN's
AR-tuned choice, not the same operator as truncation at the first \emph{greedy}
(argmax) error (Section~\ref{sec:related}). On Qwen3-8B our exact-match truncation
improves $\tau$ over the decayed-CE baseline on DFlash across all six evaluated
benchmarks, and transfers to Domino's two-branch head across the same benchmarks.
That a coarse Top-3 mask helps in an AR drafter yet hurts on a block drafter,
while exact first-error truncation helps, is itself evidence that prefix-aware
supervision does not transfer trivially from AR to mask-only drafters --- the gap
this paper is about.

On DFlash, AUF fully replaces this hand-designed position prior: once the support
is truncated at the first predicted error, applying the exponential decay
\emph{inside} that support leaves $\tau$ unchanged on all six benchmarks
(Section~\ref{sec:eval-dflash}). The position-decay weighting is thus redundant
under AUF rather than complementary to it. This is not only a simplification of
the loss: AUF removes a fixed, manually tuned inductive-bias term and lets the
model's own first-failure position decide where supervision lands --- replacing a
hand-set position prior, and its decay-rate hyperparameter $\gamma_d$, with a
model-decided support. Whether this model-decided support remains the better
default beyond the fixed settings studied here is a question we return to in
Section~\ref{sec:discussion}.

\section{Preliminaries}
\label{sec:prelim}

\paragraph{Speculative decoding.}
Let $p_\theta$ be the target model and $q_\psi$ be a draft model. At each SD
iteration, the draft model proposes a block of $B$ candidate tokens
$(\hat y_1,\ldots,\hat y_B)$. The target model verifies these candidates
left-to-right in one parallel forward pass and accepts the longest prefix that
matches the target decoding policy. Let
\[
  A = \max\{k: \hat y_i = y_i \;\; \forall i \le k\}
\]
denote the number of accepted draft tokens before the verifier emits the next
target token. In our SGLang serving path the reported acceptance length is
\(\tau=A+1\), where the extra token is this verifier-emitted correction token.
For a fixed implementation and hardware setting, higher expected \(\tau\)
directly reduces the number of target verification rounds per generated token,
so draft accuracy matters primarily through its effect on consecutive prefix
acceptance.

\paragraph{Parallel DLM/block drafters.}
DFlash is a representative block drafter: it predicts a full block in one draft
forward pass using target hidden-state features and masked block
tokens~\citep{chen2026dflash}. This removes the sequential drafting dependency
inside the block, but it creates a mismatch with SD verification. The DLM-style
drafter predicts positions in parallel, while the target verifier consumes them
strictly left-to-right. A correct prediction at position $j$ is useful only if
all previous positions are also accepted.

\paragraph{Train--inference mismatch in block objectives.}
The same mismatch can be phrased as an objective gap. DFlash-style training uses
block-level masked CE, optionally with a hand-designed position decay, while
inference measures an accepted streak under a left-to-right verifier. Thus,
training can reward later-token reconstruction even when those later tokens
would be discarded after an earlier rejection. SpecDiff-2 makes an analogous
point for diffusion drafters: diffusion models learn joint denoising
distributions over blocks, whereas autoregressive verifiers make local
prefix-conditional token decisions~\citep{sandler2025specdiff2}. It addresses
this gap by optimizing a streak-distillation objective over verifier-sampled
teacher trajectories and by adding test-time self-selection. AUF follows the
same high-level diagnosis but uses a cheaper hard-label approximation: it keeps
the existing DFlash/Domino drafter and changes only which CE terms are active.

Concretely, let
\(\ell_i(\psi)=-\log q_{\psi,i}(y_i\mid s,[\mathrm{MASK}]^{1:B})\) be the
masked-block CE at position \(i\), where all future positions are predicted in
parallel. The DFlash objective is a weighted additive reconstruction loss,
\[
  \mathcal L_{\mathrm{DFlash}}
  =
  \frac{\sum_{i=1}^{B} m_i w_i \ell_i(\psi)}
       {\sum_{i=1}^{B} m_i w_i},
  \qquad
  w_i=\exp(-(i-1)/\gamma_d),
\]
where \(m_i\) is the validity mask and \(w_i\) is the fixed decay. This loss can
emphasize early positions, but every valid suffix token remains an independent
training target. The speculative verifier instead advances by an accepted
streak,
\[
  \mathbb E[\tau\mid s]
  =
  1+\mathbb E_{\hat y_{1:B}\sim q_\psi}
  \left[\sum_{k=1}^{B}\prod_{i=1}^{k}
  \alpha_i(s,\hat y_{<i},\hat y_i)\right],
\]
where \(\alpha_i\) is the conditional acceptance probability at position \(i\)
given that the previous \(i-1\) draft tokens survived. The objective gap is
therefore structural: the decayed CE optimizes an additive per-position
surrogate, while inference rewards a product-gated prefix event.

\paragraph{A unified weighted-CE view.}
Up to per-batch normalization, the decayed CE, D-PACE's dynamic weights, and the
GRIFFIN-style prefix mask are all instances of a single weighted cross-entropy,
\begin{equation}
  \mathcal L(\psi)
  =
  \frac{\sum_{i=1}^{B} m_i\, c_i\, \ell_i(\psi)}
       {\sum_{i=1}^{B} m_i\, c_i},
  \label{eq:unified}
\end{equation}
differing only in the \emph{per-position credit} $c_i$ assigned to each CE term;
in all of them $c_i$ is detached from the gradient, performing credit assignment
without itself being optimized. Up to the constant correction-token offset in
\(\tau=A+1\), the natural smooth surrogate for accepted length is the
cumulative-confidence proxy used by
D-PACE~\citep{wu2026dpace}, $\tilde S=\sum_{k=1}^{B}\prod_{i\le k} q_{\psi,i}(y_i)$,
whose gradient decomposes into a weighted sum of per-position score gradients,
\begin{equation}
  \nabla_\psi \tilde S
  =
  \sum_{i=1}^{B} g_i\,\nabla_\psi \log q_{\psi,i}(y_i),
  \quad
  g_i = \Big(\textstyle\prod_{l\le i} q_{\psi,l}(y_l)\Big)\, f_i,
  \label{eq:gate}
\end{equation}
with continuation value $f_i=1+\sum_{m>i}\prod_{l=i+1}^{m} q_{\psi,l}(y_l)$. The
credit $g_i$ factorizes into a \emph{prefix-acceptance gate}
$\prod_{l\le i} q_{\psi,l}(y_l)$ --- a smooth estimate of the probability that the
draft prefix through position $i$ is accepted --- times a continuation value. The
gate is the dominant factor: D-PACE reports that the cumulative-only weight
recovers most of the gain over fixed decay while the continuation-only variant is
far weaker~\citep{wu2026dpace}. The reference signal is therefore that a position's
training credit should scale with the probability that its prefix is accepted, and
any objective whose credit ignores this gate is structurally misaligned with the
inference reward. SpecDiff-2 closes the gap most directly by optimizing a soft
prefix-product reward along verifier-sampled teacher continuations, at the cost of
those rollouts and a distillation
procedure~\citep{sandler2025specdiff2}; the present work instead approximates the
same gate with a hard, single-sample plug-in, developed in
Section~\ref{sec:method}.

\begin{figure}[t]
\centering
\includegraphics[width=\columnwidth]{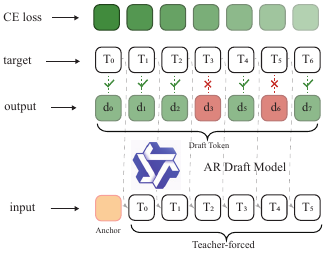}
\caption{Teacher-forced training for an autoregressive draft model. The model
receives the gold prefix as input and predicts shifted target tokens, so each CE
term is trained under the prefix-correct condition.}
\label{fig:ar-teacher-forcing}
\end{figure}

\paragraph{Teacher forcing as the source of the design.}
The reason this is natural is teacher forcing. In autoregressive training the
model predicts $y_i$ conditioned on the gold prefix $y_{<i}$, making the loss at
position $i$ a conditional-prefix loss by construction
(Figure~\ref{fig:ar-teacher-forcing}). The opposite choice --- training on
self-generated prefixes while still mapping them to gold next tokens --- creates
a distribution mismatch: once the prefix is corrupted, the supervised
input--output pair may not correspond to the data distribution~\citep{huszar2015not}.
The practical reading is that gold
supervision is the strong signal: EAGLE-3 keeps gold shifted tokens for exactly
this reason, and generates subsequent hidden states from its own training-time
rollout only because those intermediate features have no gold counterpart to
teacher-force against, not because self-generated inputs are
preferable~\citep{li2025eagle3}. The open question for a block drafter is
therefore not whether to use gold targets --- they are the strong signal --- but
which positions can legitimately receive gold supervision. This question is sharp
in speculative decoding: Domino can answer it on the \emph{input} side, because
its causal branch consumes ground-truth prefix
tokens~\citep{huang2026domino}, whereas a masked block drafter has no such channel
and must answer it on the loss side --- the route AUF takes, developed in
Section~\ref{sec:method}.

\section{Related Work}
\label{sec:related}

\paragraph{Acceptance-aware block-drafter objectives.}
The training objective we modify is shared, up to per-position weighting, across a
growing family of methods. GRIFFIN introduces a prefix mask that removes the loss
at position $j$ if an earlier target token is outside the draft model's Top-$K$
predictions, propagating the mask through a short window~\citep{hu2025griffin}.
D-PACE instead keeps the CE support intact but replaces fixed position decay with
dynamic weights derived from an accepted-length surrogate~\citep{wu2026dpace}, and
the left-to-right focal objective for diffusion drafters adds a first-error focal
term on top of a retained position decay~\citep{teachingdiffusion2026}. These three
are different treatments of the same training--verification mismatch. AUF takes the
hardest, most direct form of the signal: the drafter first emits a greedy block,
the first exact (\emph{argmax}) token mismatch defines the SD prefix boundary
$j^*$, and CE is applied only up to that boundary. The criterion and the mechanism
both differ from GRIFFIN --- an \emph{argmax} mismatch on the model's own greedy
block, not Top-$K$ membership, and a clean truncation of all later positions, not a
windowed product that can re-activate --- and from D-PACE's soft reweighting and the
focal term's additive penalty. AUF is not GRIFFIN's $K{=}1$ special case: GRIFFIN
never defines a first-error boundary or truncates past it.

Table~\ref{tab:credit} makes the relationship precise: all of these objectives are
the \emph{same} weighted block CE, differing only in the per-position credit $c_i$,
and only AUF sets $c_i$ from an exact prefix-acceptance test on the model's own
greedy block --- the hard limit of the soft prefix-acceptance gate that the other
methods grade (Section~\ref{sec:prelim}), augmented with the breaker token.

\begin{table*}[t]
\centering
\caption{Block-drafter CE objectives as instances of one weighted block CE
$\mathcal L\propto\sum_i m_i\, c_i\, \ell_i(\psi)$, differing only in the
per-position credit $c_i$ (gate defined in Section~\ref{sec:prelim}).
``Graded credit'' marks soft gates and ``verifier rollout'' marks methods that
sample teacher trajectories from $p_\theta$. Only AUF truncates the active support
to the accepted prefix plus the breaker $\{1,\ldots,j^*\}$; every prior objective
keeps the full-block support and only reweights it.}
\label{tab:credit}
\small
\setlength{\tabcolsep}{5pt}
\begin{tabular}{lccccc}
\toprule
Method & Per-position credit $c_i$ & Active support & Gate form & Graded credit & Verifier rollout \\
\midrule
Uniform CE          & $1$                                   & full block & none        & ---  & no  \\
Position decay      & $\exp(-(i-1)/\gamma_d)$               & full block & none (position prior) & ---  & no  \\
GRIFFIN Top-$K$     & $\mathbf 1[\text{prefix in Top-}K]$   & coarse mask & hard, coarse & no  & no  \\
L2R focal           & $w_i+\lambda\,\mathbf 1[i{=}j^*]$      & full block & position prior + first-error focal & no & no \\
D-PACE              & $\big(\prod_{l\le i}\tilde q_l\big) f_i$ & full block & soft       & yes & no  \\
SpecDiff-2          & product reward over $p_\theta$ paths  & full block & soft, verifier & yes & yes \\
\textbf{AUF (ours)} & $\mathbf 1[i\le j^*]$                 & $\{1,\ldots,j^*\}$ & hard + breaker & no  & no  \\
\bottomrule
\end{tabular}
\end{table*}

\paragraph{The soft-reweighting and divergence-shaping lines.}
Beyond our nearest neighbors, a broader set of recent methods keeps a full-block
support and changes how it is weighted or scored. PARD-2 reweights each position by
a cumulative product of target confidences, again replacing the fixed
decay~\citep{an2026pard2}; LK Losses optimize a direct acceptance-rate
surrogate~\citep{samarin2026lklosses} and Halfway couples drafter and target in a
joint acceptance objective~\citep{nistor2026halfway}, but both retain an exponential
position decay; and VSD lifts the objective from token likelihood to sequence-level
acceptance~\citep{zou2026vsd}. The common thread is a full-block soft support, or a
decay term left in place. AUF differs in one structural respect: it lets the
drafter's own first greedy error \emph{define} the support and then removes the
decay, rather than reweighting, augmenting, or re-shaping a support that still
spans the block.

\paragraph{Block- and diffusion-style drafters.}
The drafters we build on predict a block in parallel. DFlash is the block-diffusion
baseline whose masked-block CE with exponential decay we
target~\citep{chen2026dflash}, and Domino adds a causal branch that consumes
ground-truth prefix tokens, giving it the input-side conditioning a mask-only
drafter lacks~\citep{huang2026domino}; these are the two backbones we evaluate.
Masked-diffusion LMs can be trained as weighted masked-CE
objectives~\citep{sahoo2024simple}, and block diffusion interpolates between
autoregressive and full diffusion via the block size~\citep{arriola2025block}:
larger blocks expose more parallelism but rely more on intra-block bidirectional
aggregation. Curricula that grow the block from next-token to next-block prediction
ease adaptation from AR checkpoints and stabilize large-block
generation~\citep{tian2025nextblock,wu2026dreamreasoner}, and TraceRL argues that
DLM post-training should use information from the inference trajectory rather than
an off-policy surrogate alone~\citep{wang2025tracerl}. Together these motivate
treating AUF's active support as a training-dynamics object, not only a static loss
mask. Other recent parallel drafters scale a different axis ---
D\textsuperscript{2}SD pairs two diffusion drafters~\citep{zhang2026d2sd}, JetSpec
widens parallel tree drafting~\citep{hu2026jetspec}, and HCSpec cascades two
tiers~\citep{zhang2026hcspec} --- and are orthogonal to the training-objective
change AUF makes, which leaves the architecture and inference pipeline untouched.

\paragraph{Teacher forcing in autoregressive drafters.}
The prefix-sensitive supervision that motivates AUF is native to autoregressive
drafters, which receive the gold prefix on their input~\citep{li2025eagle3};
GRIFFIN's loss mask supplements this input-side channel rather than substituting
for it. A mask-only block drafter has no such channel, so the same intuition has
to act on the loss side (Section~\ref{sec:method}). This setting difference, not
the masking operator, is what we claim is new.

\section{Method}
\label{sec:method}

\begin{figure*}[t]
\centering
\includegraphics[width=\textwidth]{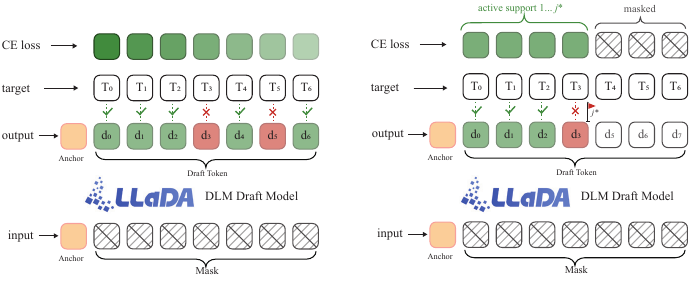}
\caption{Full-block decayed CE versus AUF for a masked block drafter. Left: the
block is predicted in parallel, and the full-block decayed CE assigns nonzero
credit to suffix tokens after the first greedy error. Right: AUF computes the
first mismatch $j^*$ from detached greedy predictions and keeps CE active only
on positions $1{:}j^*$, the accepted prefix plus the first failing token.}
\label{fig:auf-schematic}
\end{figure*}

\paragraph{Why fixed decay is train--inference inconsistent.}
The unified view of Section~\ref{sec:prelim} makes the design target explicit: a
position's credit $c_i$ in \eqref{eq:unified} should track the prefix-acceptance
gate $\prod_{l\le i} q_{\psi,l}(y_l)$ of \eqref{eq:gate}. The hand-designed
position prior sets $c_i=w_i=\exp(-(i-1)/\gamma_d)$, a function of position alone,
and is inconsistent with that gate in two ways. First, \emph{credit on
unrealizable positions}: when the prefix has already broken at some position
$j<i$, the gate is near zero but $w_i$ remains strictly positive, so the decayed CE
spends gradient raising $q_{\psi,i}(y_i)$ on positions the verifier has already
abandoned --- the block-drafter form of exposure bias. Second, \emph{a static
bottleneck}: the gate is data- and model-dependent and moves rightward as the
drafter improves, whereas $w_i$ is fixed before training, so a schedule set ex ante
cannot track the position that currently limits acceptance. Fixed decay does encode
the correct prior that earlier positions matter more --- which is why it beats
uniform CE --- but it is a position prior, not a prefix-acceptance signal, and the
two diverge exactly on the examples that determine $A$.

\paragraph{AUF as the hard plug-in of the gate.}
AUF replaces the fixed weight with the maximum-a-posteriori, single-sample plug-in
of the same gate. Replacing each soft factor $q_{\psi,l}(y_l)$ in the prefix
product by its greedy realization $\mathbf 1[\hat y_l=y_l]$ collapses the gate to
the indicator $\prod_{l\le i}\mathbf 1[\hat y_l=y_l]=\mathbf 1[i< j^*]$, where
$j^*$ is the first greedy mismatch. This hard gate activates only the accepted
prefix $\{1,\ldots,j^*{-}1\}$; AUF extends the support by one position to include
the failing token $j^*$ itself, because it is the decision that breaks the accepted
prefix and the position whose improvement directly extends $A$. The resulting
support $\{1,\ldots,j^*\}$ is therefore not the gate alone but the gate plus the
breaker --- the minimal superset that covers every position relevant to the next
unit of acceptance. AUF is thus a zero-temperature, hard version of the
prefix-acceptance gate that D-PACE estimates softly, augmented with the single
token whose CE gradient most directly pushes $j^*$ rightward, applied as a $0/1$
support instead of a graded weight. Unlike fixed decay it retargets automatically:
as the drafter improves, $j^*$ moves right and the active support expands with no
schedule. We give the loss next, then return to what this hard plug-in gives up.

\paragraph{Accept-Until-Fail CE.}
For a training block of length $B$, let $V=\{i:m_i=1\}$ be the set of natively
valid positions (the validity mask of \eqref{eq:unified}), let $y_i$ be the target
token and $\hat y_i=\arg\max q_\psi(\cdot)_i$ be the draft model's current greedy
token at position $i$ (Figure~\ref{fig:auf-schematic}). AUF finds the first exact
mismatch among valid positions,
\[
  j^* = \min\{j\in V: \hat y_j \ne y_j\}.
\]
If such a position exists, the active CE support is
$S=\{i\in V:i\le j^*\}$; otherwise $S=V$. The loss is
\[
  \mathcal{L}_{\mathrm{AUF}}
  =
  \frac{1}{|S|}\sum_{i\in S} -\log q_\psi(y_i)_i.
\]
The first failing token is kept in the support because it is the decision that
breaks the accepted prefix. Tokens after the first failure are inactive because
they would not be verified as accepted draft tokens in the corresponding SD
iteration. Equivalently, AUF replaces the fixed DFlash weights $w_i$ with the
model-dependent hard prefix weight $a_i(\psi)=\mathbf 1[i\le j^*]$ (with $a_i=1$
for all $i$ on a fully correct block), so the loss is the weighted CE
\eqref{eq:unified} with $c_i=a_i(\psi)$. The mask $a_i(\psi)$ is computed from
detached predictions and is not a gradient path: AUF remains a CE objective whose
only change is the credit-assignment rule over positions.

\paragraph{What AUF gives up, and what it costs.}
Relative to the soft gate of \eqref{eq:gate}, the hard plug-in discards two
things. First, it removes graded credit: a near-miss at the breaker and a
confidently correct early token both receive credit $1$, whereas the soft gate
down-weights a fragile prefix. Second, a single deterministic argmax is a coarse,
$0/1$ estimator of the gate: on any given example the credit is binary rather than
graded, so per-example information about prefix confidence is lost. These are
exactly the two properties that D-PACE's
smoothing and SpecDiff-2's product reward preserve at extra cost; AUF trades that
graded information for a parameter-free hard support. What AUF buys in return is cost, simplicity, and one fewer inductive
bias: it changes only the coefficient $c_i$ in the existing decayed-CE loss, needs
only a detached greedy readout from the training logits to locate $j^*$, adds no
auxiliary objective, samples no verifier trajectories, removes the
decay hyperparameter $\gamma_d$, and leaves the inference pipeline and the
rejection-sampling exactness contract untouched.

\paragraph{Two optimization axes.}
AUF changes training along two measurable axes. First, it changes the
\emph{active token ratio}
\[
  r_{\mathrm{active}} = |S|/B,
\]
which measures how much of the block participates in the CE update: early in
training the first error appears near the beginning, so $r_{\mathrm{active}}$ is
small, and as the drafter pushes the first failure rightward the support expands.
Second, it induces the diagnostic \emph{support-token accuracy}
\[
  a_{\mathrm{support}}
  =
  \frac{1}{|S|}\sum_{i\in S}\mathbf{1}[\hat y_i=y_i],
\]
measured on the same first-error support that AUF selects for CE. For AUF this is
not an independent fixed-denominator accuracy: because the support is the accepted
prefix plus the breaker token, it is tied to the current first-failure depth. We
use it as a training-side diagnostic of where the support lies, and use
common-token accuracy and inference-side conditional acceptance for
fixed-denominator comparisons.

\begin{figure}[t]
\centering
\includegraphics[width=\columnwidth]{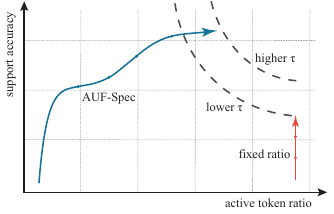}
\caption{A two-dimensional training diagnostic over training steps, drawn
schematically from the per-step measurements of
Figure~\ref{fig:dflash-validation}. Axes: $x=$ active token ratio
$r_{\mathrm{active}}=|S|/B$, $y=$ accuracy on
the selected first-error support $a_{\mathrm{support}}$. Dashed contours
schematically indicate higher acceptance length $\tau$ toward the upper right.
A fixed-support method moves almost vertically (constant $r_{\mathrm{active}}$,
accuracy measured on the same positions). AUF instead traces an
expanding-support path where the active support grows as the first failure $j^*$
moves rightward.}
\label{fig:two-axis}
\end{figure}

This exposes an optimization dimension that fixed-support objectives --- uniform
CE, fixed decay, and dynamic reweighting such as D-PACE --- do not have: they keep
the supervised token set fixed and improve only the accuracy or weighting of the
same positions, whereas AUF can also improve by pushing the first failure later,
growing $r_{\mathrm{active}}$. AUF thus turns the left-to-right acceptance boundary
itself into part of the training signal. We read the resulting trajectory as an
implicit easy-to-hard curriculum the drafter sets for itself: early on it is
supervised to be reliable on a short accepted prefix (small $r_{\mathrm{active}}$),
and as that prefix competence grows it pushes $j^*$ rightward and extends the
supervised horizon with it, so a single driver moves both diagnostics up and to
the right rather than two independently scheduled quantities
(Figure~\ref{fig:two-axis}). The native serving block size stays fixed; only the
loss horizon grows, per example and per step, from the drafter's realized prefix
rather than a manually fixed block-size schedule --- mirroring the block-diffusion
observation that fine-grained local correctness should be established before
exploiting large parallel blocks. This is an interpretation of the training
dynamics measured in Section~\ref{sec:eval-dflash}, not a separate mechanism.

\paragraph{AUF as a teacher-forcing-motivated support rule.}
Stated as interpretation rather than theorem, AUF is the loss-side realization of
the teacher-forcing idea of Section~\ref{sec:prelim} for drafters without an
input-side gold-prefix channel. The acceptance-aware methods push back toward
prefix-sensitive supervision but mostly by reweighting a full-block loss whose
support still spans the block; each changes \emph{how much} a position counts
without changing \emph{which} positions receive CE. The first-error focal
objective even keys on the same breaker $j^*$ yet \emph{adds} a term on top of the
full decayed CE, leaving the suffix supervised, whereas AUF sets $c_i=0$ for
$i>j^*$ and \emph{removes} it; among drafters, only Domino changes the
conditioning event on the input side. The analogy is intentionally limited:
strict teacher forcing conditions on a \emph{counterfactual} gold prefix, whereas
AUF conditions the \emph{support} on the drafter's own realized greedy prefix up
to $j^*$ while the \emph{target} at every supervised position stays the gold
token. We therefore use teacher-forcing language as motivation, not as a claim
that AUF exactly recovers the same conditional.

\paragraph{Applying AUF to different drafters.}
For DFlash, AUF is applied directly to the block logits. For Domino, which has a
base branch and a final correction branch, the first-error support is computed
separately for each output distribution. The change is scoped to the CE support
and modifies neither the drafter architecture nor the inference-time verifier
contract.

\section{Evaluation}
\label{sec:evaluation}

\paragraph{Setup.}
We evaluate Qwen3-8B drafters trained on ShareGPT. The target model is
\texttt{Qwen3-8B}. We use the original ShareGPT training conversations directly,
without regenerating assistant responses from the target model. For Qwen3 we use
the standard \texttt{qwen} chat template rather than the thinking-enabled
\texttt{qwen3-thinking} template, so thinking is disabled both when formatting
training data and when rendering evaluation prompts. We treat thinking-enabled
Qwen3 as a different serving setting and do not vary it here. DFlash uses a
5-layer block drafter with target layers $\{1,9,17,25,33\}$ and block size
$B=16$; Domino uses the same backbone with its GRU-prefix base/final two-branch
head. Our baseline is the standard decayed-CE configuration ($\gamma=7$)
recommended by the SpecForge training recipe~\citep{chen2026dflash,huang2026domino};
this is the default used in both DFlash and Domino and reproduces the exponential
position decay adopted across the block-drafter literature (EAGLE-3, DFlash,
Domino all hand-set an exponential $\gamma$~\citep{li2025eagle3,chen2026dflash,huang2026domino}).
AUF keeps the architecture and training data fixed and only changes the supervised
support, replacing the full-block decayed CE support with the first-error support
of Section~\ref{sec:method}. All decode numbers are produced by our SGLang
DFlash/Domino V2 speculative-decoding backend (single GPU, batch size 1) under
two verification settings: greedy ($T{=}0$) and sampling ($T{=}1$); in both
cases the draft proposal is greedy (single top-1 candidate per position).
Because every comparison in this section is \emph{within} a fixed drafter
architecture (AUF is a training recipe, not a new model), per-iteration forward
cost is identical across the variants we compare. We therefore report average
acceptance length $\tau$ (mean accepted draft tokens plus one correction token
per speculative iteration, as reported by SGLang's built-in metrics) as the
main decode metric throughout this section.

\paragraph{Benchmarks.}
Both backbones are evaluated on the same six benchmarks: GSM8K, MATH-500,
HumanEval, MBPP, MT-Bench, and Alpaca, with prompt counts 200, 200, 164, 200,
80, 200 and a uniform \texttt{max\_new\_tokens=512} cap across all methods and
temperatures. Every drafter is trained for six epochs on ShareGPT; the decode
tables report the epoch-6 checkpoints, and the training-dynamics figures use the
full epoch-1 optimizer-step logs. We separate the two backbones into their own
subsections because the Domino two-branch head admits a richer family of AUF
variants.

\subsection{DFlash}
\label{sec:eval-dflash}

DFlash has a single output distribution, so the AUF design space is small. We
train three matched variants: the official-style \emph{decay-only} reference
($\gamma=7$ full-block CE), \emph{AUF-only} (first-error truncated CE, no decay),
and \emph{AUF+decay} (the same exponential decay applied \emph{inside} the AUF
support). Table~\ref{tab:dflash-main} reports the epoch-6 decode results.

\begin{table*}[t]
    \caption{DFlash epoch-6 average acceptance length ($\tau$) on Qwen3-8B across six benchmarks, measured through the SGLang serving path (batch size 1). Best per temperature setting in bold.}
    \label{tab:dflash-main}
    \centering
    \small
    \setlength{\tabcolsep}{5pt}
    \newlength{\benchwidth}
    \settowidth{\benchwidth}{\small MATH-500\hspace{2pt}}
    \begin{tabular}{c @{\hspace{0.6em}} c @{\hspace{1.0em}} w{c}{\benchwidth} w{c}{\benchwidth} @{\hspace{1.0em}} w{c}{\benchwidth} w{c}{\benchwidth} @{\hspace{1.0em}} w{c}{\benchwidth} w{c}{\benchwidth} @{\hspace{1.0em}} w{c}{\benchwidth}}
        \toprule
        \multirow{2}[2]{*}{Temperature} & \multirow{2}[2]{*}{Method}
        & \multicolumn{2}{c@{\hspace{1.0em}}}{\sc{Math}}
        & \multicolumn{2}{c@{\hspace{1.0em}}}{\sc{Code}}
        & \multicolumn{2}{c@{\hspace{1.0em}}}{\sc{Chat}}
        & {\sc{Overall}} \\
        \cmidrule(lr){3-4}
        \cmidrule(lr){5-6}
        \cmidrule(lr){7-8}
        \cmidrule(lr){9-9}
        & & GSM8K & MATH-500 & HumanEval & MBPP & MT-Bench & Alpaca & \textit{Avg.} \\
        \midrule
        \multirow{3}{*}{$T{=}0$}
        & Decay-only & 2.29 & 2.39 & 2.86 & 2.85 & 2.03 & 2.01 & 2.40 \\
        & AUF+decay & 2.50 & \textbf{2.64} & \textbf{3.09} & 3.08 & \textbf{2.19} & 2.14 & \textbf{2.61} \\
        & AUF-only & \textbf{2.51} & \textbf{2.64} & \textbf{3.09} & \textbf{3.09} & 2.18 & \textbf{2.15} & \textbf{2.61} \\
        \midrule
        \multirow{3}{*}{$T{=}1$}
        & Decay-only & 2.21 & 2.28 & 2.70 & 2.74 & 1.94 & 1.96 & 2.31 \\
        & AUF+decay & 2.39 & 2.49 & 2.88 & 2.93 & \textbf{2.08} & 2.09 & \textbf{2.48} \\
        & AUF-only & \textbf{2.40} & \textbf{2.50} & \textbf{2.89} & \textbf{2.94} & 2.06 & \textbf{2.11} & \textbf{2.48} \\
        \bottomrule
    \end{tabular}
\end{table*}

\begin{figure*}[t]
\centering
\includegraphics[width=0.9\textwidth]{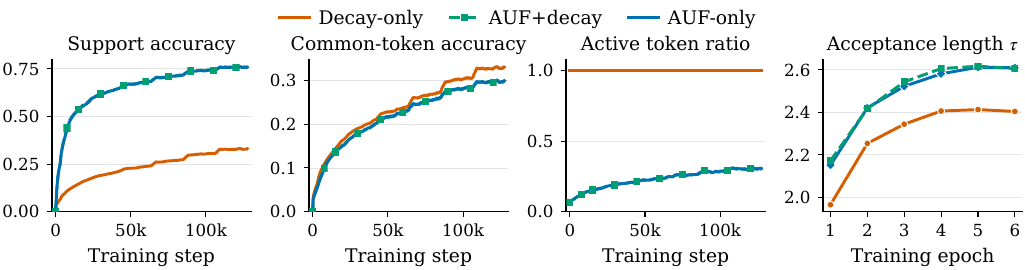}
\caption{DFlash training dynamics across epochs (rolling mean over $3000$
steps; $\tau$ from six-benchmark greedy SGLang decode). Panels:
\textbf{(1)} support accuracy, on the positions each method actually trains on
(for AUF, the selected first-error support);
\textbf{(2)} common-token accuracy, on the fixed common mask of all
natively-valid block positions; \textbf{(3)} active token ratio, the fraction
of block positions each method supervises; \textbf{(4)} acceptance length
$\tau$. See text for interpretation.}
\label{fig:dflash-validation}
\end{figure*}

Both AUF variants improve over the decay-only reference on every benchmark under
both greedy and sampling decoding (Table~\ref{tab:dflash-main}), with the largest
gains on the longer, more structured code and math outputs; the sampling result
confirms the gain is not an artifact of the greedy verification match used during
training, and the advantage holds throughout training, not just at convergence
(Figure~\ref{fig:dflash-validation}). Notably, \emph{AUF-only and AUF+decay are
indistinguishable}: once the support is truncated at the first predicted error,
re-weighting the surviving positions with exponential decay adds nothing
measurable. We read this decay-inert result in Section~\ref{sec:discussion}.

\paragraph{Training dynamics.}
Figure~\ref{fig:dflash-validation} logs two token accuracies per optimizer step:
\emph{common-token accuracy}, measured on a fixed common mask (all natively-valid
block positions) so the three methods share a denominator, and the
\emph{support accuracy} of Section~\ref{sec:method}, measured on the positions each
method actually trains on (for AUF, the selected first-error support). Two facts
stand out. First, on the common mask the \emph{decay-only} baseline is the
\emph{most} accurate, yet it is the \emph{worst} at decode. Second, AUF's active
token ratio starts low and grows over training while decay-only trains the full
block, so the AUF support diagnostic tracks a much shorter first-error region.
This common-mask accuracy inversion --- more accurate yet slower --- is the
empirical signature we interpret in Section~\ref{sec:discussion}.

\begin{figure}[t]
\centering
\includegraphics[width=\columnwidth]{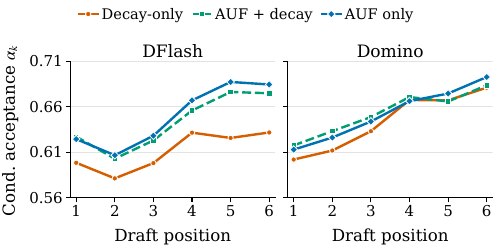}
\caption{Per-position conditional acceptance $\alpha_k$ (epoch~6, greedy SGLang
decode), averaged over all six benchmarks. Aggregation weights prompts equally
within each benchmark and then weights the six benchmarks equally; MT-Bench's two
turns count as separate prompts. Left: DFlash decay-only, AUF+decay, and
AUF-only. Right: Domino decay-only and the two base-branch AUF variants
(B-AUF+D, B-AUF), using the same line styles. AUF raises the six-benchmark
conditional acceptance profile on both backbones.}
\label{fig:cond-acceptance}
\end{figure}

\paragraph{Per-position conditional acceptance.}
The matching inference-side measurement is the per-position \emph{conditional}
acceptance rate $\alpha_k$ --- among speculative steps whose draft prefix through
position $k-1$ was fully accepted, the fraction that also accept position $k$,
\[
  \alpha_k
  =
  \frac{\#\{A \ge k\}}{\#\{A \ge k-1\}},
\]
read off the accepted-draft-token histogram the SGLang verifier records per
request. This is exactly the per-position factor in
$\mathbb E[\tau]=1+\sum_k\prod_{i\le k}\alpha_i$ (Section~\ref{sec:prelim}): it
removes the compounding penalty of earlier rejections and isolates conditional
predictive quality at each block offset. We plot positions $1$--$6$ only, where
every curve retains a non-trivial fraction of speculative steps in the
conditioning set under the six-benchmark average; deeper positions are too thin
to be stable. On DFlash (Figure~\ref{fig:cond-acceptance}, left) both AUF variants lie
above the decay-only baseline across the block, and AUF-only and AUF+decay are
again nearly indistinguishable: the decode gain is a lift of the conditional
acceptance profile that determines $\tau$, not a global shift in raw token
accuracy.

\subsection{Domino}
\label{sec:eval-domino}

Domino is harder to analyze because its head is not a single distribution: it
mixes a \emph{base} branch (the drafter's own logits) and a \emph{final} branch
(base plus a causal correction), trained as
$L=\lambda_{\mathrm{base}}L_{\mathrm{base}}+(1-\lambda_{\mathrm{base}})L_{\mathrm{final}}$.
AUF therefore has a choice of \emph{which support each branch trains on}, and
there is no a-priori ``main'' AUF line --- we treat the branch-support assignment
as an experimental variable and let the measured decode decide. We evaluate seven
trained variants spanning that space (Table~\ref{tab:domino-variants}): three
support assignments (base-only, base-shared, branch-specific), each with and
without the $\gamma=7$ decay applied inside the AUF support, plus the decay-only
baseline.

\begin{table}[t]
\centering
\caption{Domino AUF variants. ``Support'' is the training mask used by each branch: \texttt{common} = full native block, \texttt{base-auf}/\texttt{final-auf} = first-error truncation from the base/final branch. Decay is the $\gamma=7$ position weight applied inside the support. Short names are used in subsequent tables.}
\label{tab:domino-variants}
\small
\setlength{\tabcolsep}{4pt}
\begin{tabular}{llll}
\toprule
Short name & base support & final support & decay \\
\midrule
Decay-only (baseline) & common & common & on \\
\midrule
B-AUF & base-auf & common & off \\
B-AUF+D & base-auf & common & on \\
S-AUF & base-auf & base-auf & off \\
S-AUF+D & base-auf & base-auf & on \\
F-AUF & base-auf & final-auf & off \\
F-AUF+D & base-auf & final-auf & on \\
\bottomrule
\end{tabular}
\vspace{0.3em}

{\footnotesize \textbf{B} = base-only (AUF on base branch only); \textbf{S} = shared (both branches share base-auf support); \textbf{F} = branch-specific (final branch uses its own final-auf support). \textbf{+D} = decay applied inside the AUF support.}
\end{table}

\begin{table*}[t]
    \caption{Domino epoch-6 average acceptance length ($\tau$) on Qwen3-8B across six benchmarks, same SGLang protocol as Table~\ref{tab:dflash-main}. Best per temperature block in bold; all AUF variants beat the decay-only baseline. Short names refer to Table~\ref{tab:domino-variants}.}
    \label{tab:domino-main}
    \centering
    \small
    \setlength{\tabcolsep}{5pt}
    \settowidth{\benchwidth}{\small MATH-500\hspace{2pt}}
    \begin{tabular}{c @{\hspace{0.6em}} l @{\hspace{1.0em}} w{c}{\benchwidth} w{c}{\benchwidth} @{\hspace{1.0em}} w{c}{\benchwidth} w{c}{\benchwidth} @{\hspace{1.0em}} w{c}{\benchwidth} w{c}{\benchwidth} @{\hspace{1.0em}} w{c}{\benchwidth}}
        \toprule
        \multirow{2}[2]{*}{Temperature} & \multirow{2}[2]{*}{Method}
        & \multicolumn{2}{c@{\hspace{1.0em}}}{\sc{Math}}
        & \multicolumn{2}{c@{\hspace{1.0em}}}{\sc{Code}}
        & \multicolumn{2}{c@{\hspace{1.0em}}}{\sc{Chat}}
        & {\sc{Overall}} \\
        \cmidrule(lr){3-4}
        \cmidrule(lr){5-6}
        \cmidrule(lr){7-8}
        \cmidrule(lr){9-9}
        & & GSM8K & MATH-500 & HumanEval & MBPP & MT-Bench & Alpaca & \textit{Avg.} \\
        \midrule
        \multirow{7}{*}{$T{=}0$}
        & Decay-only & 2.42 & 2.52 & 3.10 & 3.08 & 2.13 & 2.13 & 2.56 \\
        \cmidrule(l){2-9}
        & B-AUF & 2.51 & 2.65 & \textbf{3.21} & 3.17 & 2.23 & 2.20 & 2.66 \\
        & B-AUF+D & 2.52 & 2.65 & \textbf{3.21} & \textbf{3.18} & \textbf{2.26} & \textbf{2.23} & \textbf{2.68} \\
        & S-AUF & 2.51 & 2.66 & 3.15 & 3.11 & 2.20 & 2.17 & 2.63 \\
        & S-AUF+D & 2.52 & 2.66 & 3.17 & 3.11 & 2.23 & 2.18 & 2.65 \\
        & F-AUF & 2.49 & 2.63 & 3.16 & 3.09 & 2.21 & 2.18 & 2.63 \\
        & F-AUF+D & \textbf{2.54} & \textbf{2.67} & 3.14 & 3.15 & 2.23 & 2.18 & 2.65 \\
        \midrule
        \multirow{7}{*}{$T{=}1$}
        & Decay-only & 2.33 & 2.38 & 2.89 & 2.94 & 2.02 & 2.11 & 2.44 \\
        \cmidrule(l){2-9}
        & B-AUF & 2.41 & \textbf{2.52} & 2.98 & \textbf{3.04} & \textbf{2.11} & 2.12 & \textbf{2.53} \\
        & B-AUF+D & 2.40 & 2.49 & \textbf{3.03} & 3.01 & 2.08 & \textbf{2.14} & 2.52 \\
        & S-AUF & \textbf{2.45} & 2.49 & 2.94 & 2.96 & 2.07 & 2.13 & 2.51 \\
        & S-AUF+D & 2.40 & 2.50 & 2.95 & 2.97 & 2.07 & 2.12 & 2.50 \\
        & F-AUF & 2.40 & 2.47 & 2.99 & 2.99 & 2.06 & 2.13 & 2.51 \\
        & F-AUF+D & \textbf{2.45} & 2.48 & 2.93 & 2.99 & 2.08 & 2.09 & 2.50 \\
        \bottomrule
    \end{tabular}
\end{table*}

Table~\ref{tab:domino-main} shows the support-selection idea transfers: every
AUF variant beats the decay-only baseline under both greedy and sampling
decoding, regardless of how the two branches split their support. Three
observations sharpen the picture.

First, the empirically strongest variant under greedy decoding is B-AUF+D, which
applies AUF to the base branch, leaves the \emph{final} branch on the common mask,
and retains the decay inside the AUF support; under sampling, B-AUF (without decay)
is marginally stronger. The gain is localized to the \emph{base} (proposer) branch:
its training loss drops sharply once AUF is applied, while pushing AUF onto the
final branch as well does not help (and slightly hurts) --- a degree of freedom
DFlash's single distribution does not expose.

Second, the spread among the six AUF variants is small, so we report the ordering
rather than claim a designed main line. The smaller absolute Domino margin than on
DFlash is consistent with its stronger baseline: a higher starting point leaves
less headroom for support reshaping.

Third, unlike DFlash where decay on top of AUF is inert
(Table~\ref{tab:dflash-main}), on Domino decay adds a small benefit under greedy
decoding across all three matched decay-toggle pairs that vanishes or reverses
under sampling. The residual is modest and regime-dependent; we take up whether it
carries information beyond the support selection in Section~\ref{sec:discussion}.

\paragraph{Per-position conditional acceptance on Domino.}
Domino's two-branch head has no single training-accuracy axis to plot, so we read
the effect of AUF off the inference side, with the same $\alpha_k$ used for DFlash
(Figure~\ref{fig:cond-acceptance}, right). The two base-branch AUF variants
(B-AUF and B-AUF+D) lift the conditional acceptance profile over decay-only
through most of the block, but by a smaller margin than on DFlash, consistent with
Domino's stronger baseline and narrow $\tau$ spread.

\section{Discussion}
\label{sec:discussion}

\paragraph{Why does AUF help?}
AUF improves $\tau$ over the decayed-CE baseline across our runs
(Section~\ref{sec:evaluation}). We are confident about the method's structure and
its train--inference-consistency motivation (Section~\ref{sec:method}), but not
yet about the precise causal mechanism, which likely needs targeted
interpretability analysis we leave to future work. The principled explanation is
that the full-block CE supervises positions whose prefix the verifier never
reaches, learning a target less prefix-sensitive than inference itself, whereas
AUF truncates the support at the first predicted failure and thus optimizes a
quantity that tracks SGLang's acceptance length $\tau$ more closely. This is
principled but not proven: ``better aligned objective'' and ``higher measured
$\tau$'' are not the same statement, and we treat the causal link as a hypothesis.

Two readings refine this. The first is structural: AUF is a projection of an RL
objective onto a supervised one. Maximizing the verifier's expected acceptance
length $\mathbb E[\tau]$ would sample rollouts from the current policy, assign
credit along them, and update with a reward-weighted gradient; AUF keeps the
on-policy rollout (the greedy block) and the acceptance-aligned credit rule (the
first-error position $j^*$ deciding \emph{where} to supervise) but replaces the
reward-weighted policy gradient with the ordinary gold CE
$-\sum_{i\le j^*}\nabla_\psi\log q_{\psi,i}(y_i)$. This explains the appeal --- a
stable dense CE gradient that still carries acceptance-aligned credit --- without
our having measured that this is \emph{why} it wins. The second is a falsifiable
prediction: supervising post-failure positions is an off-deployment task that
competes for capacity with the early positions determining the accepted prefix,
so fixed-support baselines should buy higher generic token accuracy at the expense
of decode. This is exactly the inversion DFlash shows
(Section~\ref{sec:eval-dflash}); what it does not settle is whether the effect is
causal data-curation or the same inversion the consistency reading already
predicts.

\paragraph{A spectrum, with AUF as the minimal intervention.}
The existing objectives lie on one axis from supervised imitation toward
verifier-calibrated RL: uniform CE $\to$ position decay $\to$ AUF $\to$ D-PACE
$\to$ SpecDiff-2. Uniform CE and fixed decay encode a position prior with no
acceptance signal; AUF is the first point that conditions on the model's realized
prefix, using a hard on-policy support and a gold target; D-PACE softens the same
gate into a graded weight; SpecDiff-2 moves the target onto verifier trajectories,
approaching the RL objective itself~\citep{sandler2025specdiff2}. AUF is thus the
\emph{minimal} edit that conditions training on the realized accepted prefix, and
is distinct from sequence-level rejection-sampling fine-tuning: it performs
block-internal, token-prefix-level credit assignment tied to the left-to-right
acceptance semantics of speculative decoding.

\paragraph{One fewer inductive bias.}
A practical consequence is that AUF removes a hand-designed component rather than
adding one: it replaces the exponential position decay --- a fixed, manually tuned
weight --- with the drafter's own first-failure boundary, which the DFlash result
shows carries the same ``earlier tokens matter more'' prior \emph{and} the
acceptance signal the fixed prior lacks. This removes the decay rate $\gamma_d$
from the hyperparameter budget, and because the support is computed from the model
rather than fixed ex ante, the rule is defined identically regardless of block
size $B$ or backbone. We stress this is a structural motivation, not an empirical
scaling claim.

\paragraph{Does AUF replace decay, or only supplement it?}
On DFlash this is answered cleanly: AUF-only and AUF+decay are indistinguishable
under both greedy and sampling decoding (Section~\ref{sec:eval-dflash}), so decay
is fully redundant once the support is selected. On Domino a small greedy-regime
residual survives (B-AUF+D $2.68$ vs.\ B-AUF $2.66$) but reverses under sampling
(B-AUF $2.53$ vs.\ B-AUF+D $2.52$); averaged across regimes the decay-on and
decay-off variants are within noise and the sign flips with temperature. We
therefore conclude that decay is not a necessary complement to AUF on either
backbone.

\paragraph{Limitations.}
All results are at a single configuration: block size $B=16$, target Qwen3-8B, and
ShareGPT training data. Three axes remain unverified because each requires a full
drafter retraining run. We do not vary $B$ or model scale, so the broader-default
argument above is a structural expectation, not a confirmation; we do not include
head-to-head retraining against D-PACE or other acceptance-aware baselines under
matched compute; and whether AUF's stable SFT-side update is preferable to a tuned
on-policy RL objective, or only cheaper, is untested.

\bibliography{references}

\end{document}